\documentclass{article}

\usepackage{arxiv}

\usepackage[utf8]{inputenc} 
\usepackage[T1]{fontenc}    
\usepackage{hyperref}       
\usepackage{url}            
\usepackage{booktabs}       
\usepackage{amsfonts}       
\usepackage{nicefrac}       
\usepackage{microtype}      
\usepackage{lipsum}		
\usepackage{graphicx}
\usepackage[square,numbers]{natbib}
\usepackage{doi}
\usepackage{subfigure}

\usepackage{arydshln}

\title{Neural Network Compression for Reinforcement Learning Tasks}

\author{Dmitry A. Ivanov \\
	Lomonosov Moscow State University\\
	Moscow, Russia \\
        Institute of Applied Physics of the R.A.S.,\\
        Nizhny Novgorod, Russia \\
	\texttt{rudimiv@gmail.com} \\
	\And
	Denis A. Larionov \\
	Chuvash State University\\
        Cheboksary, Russia \\
        Cifrum \\
	Moscow, Russia \\
        \And
        Oleg V. Maslennikov \\
        Institute of Applied Physics of the R.A.S.,\\
        Nizhny Novgorod, Russia \\
        \And
        Vladimir V. Voevodin \\
	Lomonosov Moscow State University\\
	Moscow, Russia \\
}

\renewcommand{\shorttitle}{\textit{arXiv} Template}

\begin{document}
\maketitle

\begin{abstract}
    In real applications of Reinforcement Learning (RL), such as robotics, low latency and energy efficient inference is very desired. The use of sparsity and pruning for optimizing Neural Network inference, and particularly to improve  energy and latency efficiency, is a standard technique. 
    In this work, we perform a systematic investigation of applying these optimization techniques for different RL algorithms in different RL environments, yielding up to a 400-fold reduction in the size of neural networks.
\end{abstract}


\section{Introduction}


In the last decade, neural networks (NNs) have driven significant progress across various fields, notably in deep reinforcement learning, highlighted by studies like \cite{mnih2015human, rl_tokamak, rl_drone}. This progress has the potential to make changes in many areas such as embedded devices, IoT and Robotics.
Although modern Deep Learning models have demonstrated impressive gains in accuracy, their large sizes pose limits to their practical use in many real-world applications \cite{quant_survey}. These applications may impose requirements in energy consumption, inference latency, inference throughput, memory footprint, real-time inference and hardware costs. 


Numerous studies have attempted to make neural networks more efficient. These approaches can generally be categorized at least into the next several groups \cite{quant_survey}: pruning \cite{liang2021pruning}, temporal sparsity \cite{yousefzadeh2019asynchronous}, \cite{significant_multiplications}, distillation \cite{hinton2015distilling}, quantization \cite{liang2021pruning}, neural architecture search of efficient NN architectures \cite{ren2021comprehensive} , hardware and NN co-design \cite{eie}. Additionally, some works try to mix some of these methods \cite{deep_compression}. The combination of methods could lead to substantial improvements in neural network efficiency. E.g. the combination of 8-bit integer quantization and 10\% sparsity may result in a 40x times decrease in memory footprint and a decrease in computational complexity achieved by using fewer arithmetical operations and using integer arithmetic. Furthermore, beyond the efficiency gains, the introduction of sparsity may contribute to enhanced accuracy in neural networks. For example, it was shown in \cite{blalock2020state, rl_sparse, significant_multiplications} that sparse neural networks derived by pruning usually achieve better results than their dense counterparts with an equivalent number of parameters. Moreover, even sparse neural networks that contain 10\% of the weights of the original network sometimes could achieve higher accuracy than dense neural networks \cite{frankle2018lottery}. 


Several of the previously mentioned NN optimization methods find inspiration in neuroscience. In the brain, the presence of dense layers is not evident. Instead, the brain employs a mechanism akin to rewiring and pruning to eliminate unnecessary synapses \cite{principles_of_ns}. As a result, brain neural networks are sparse and have irregular topology. Several works in neuroscience state that the brain represents and processes information in the discrete/quantized form \cite{brain_discrete}, \cite{perception_discrete}. It could be justified that information stored in continuous form would be inevitably corrupted by noise that is present in any physical system \cite{noise_nervous_system}. It is impossible to measure a physical variable with infinite precision. Moreover, from the point of view of the Bayesian framework quantization leads to stability in the representation of information and robustness to additive noise \cite{bayesian_framework}. 

\begin{figure}
  \includegraphics[width=0.8\linewidth]{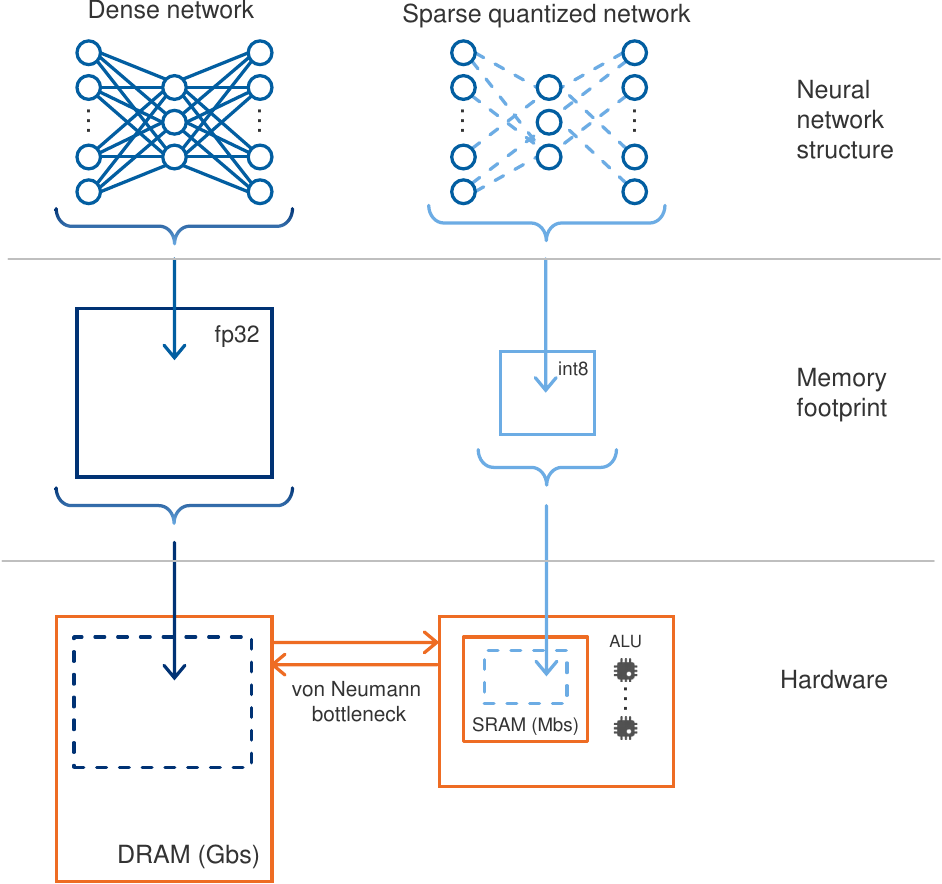}
  \caption{Illustration of the fitting of dense NN to DRAM memory and sparse and quantized NN to SRAM memory.}
  \label{fig:sram}
\end{figure}

It is well known that obtaining data from Dynamic Random Access Memory (DRAM) is much more expensive in terms of energy and time in comparison to arithmetic operations and obtaining data from fast, but expensive Static Random Access Memory (SRAM) \cite{horowitz2014}. This problem is commonly known as the von Neumann problem \cite{backus1978can, neuromorphic_systems}. The huge sizes of contemporary neural networks exacerbate this problem, making it difficult to achieve high Frames Per Second (FPS), low latency, and energy-efficient performance on modern hardware. The reduction of neural network sizes leads to the diminishing of data exchange between memory and processor and potentially results in higher performance and less energy consumption. Furthermore, a significant reduction in neural network size may enable its placement in faster SRAM, contributing to a notable improvement in memory access latency and throughput (see Fig. \ref{fig:sram}). It is worth noting that reducing memory accesses is much more significant for speeding up neural networks than just the reduction of arithmetic operations \cite{eie}.

However, there are only a few papers \cite{quarl, rl_sparse} that apply these approaches to RL. And to the best of our knowledge, there are no papers that try to mix them. At the same time, many potential RL applications impose strong latency, FPS and energy limits. For example, in \cite{rl_tokamak} DeepMind applied RL for tokamak control, and it was necessary to achieve a remarkable 10kHz FPS to meet the operational requirements. Similarly, in \cite{rl_drone} authors applied RL for drone racing. Drone control requires 100 HZ FPS for the RL network. Moreover, since the network inference was on board, strong restrictions are placed on energy consumption. These examples highlight the critical need for advancing optimization techniques to  RL to meet the demanding performance criteria of various applications.

In this work, we apply a combination of quantization and pruning techniques for RL tasks. The primary goals were to showcase the possibility of dramatically improving the efficiency of actor networks trained using various RL algorithms and to investigate the applicability of NN optimization techniques and their combination in the RL context. The ultimate aim is to broaden the applicability of these networks to a diverse array of embedded applications, particularly those with strong requirements for FPS, energy consumption, and hardware costs. Our findings indicate that it is feasible to apply quantization and pruning to Neural Networks trained by RL without loss in accuracy. Furthermore, sometimes we even observed improvements in accuracy after applying these optimization techniques. This suggests a promising avenue for optimizing RL-based actor networks for resource-constrained embedded applications without sacrificing performance.

\section{Background}
\subsection{RL}

In RL \cite{sutton2018reinforcement}, an agent interacts with an environment by sequentially selecting actions $a$ in response to the current environment state $s$. After making a choice, it transits to a new state $s'$ and receives a reward $r$. The agent's goal is to maximize the sum of discounted rewards.

This is formalized as a Markov decision process defined as a tuple $(S, A, P, R)$, where $S$ is the set of states, and $A$ is the set of actions. $P$ is the function describing transition between states; $P(s' | s, a) = Pr(s_{t+1} = s' | s_t = s, a_t = a)$, \textit{i.e.}, is the probability to get into state $s'$ at the next step when selecting action $a$ in state $s$. $R = R(s, a, s')$ is the reward function that determines the reward an agent will receive when transitioning from state $s$ to state $s'$ by selecting an action $a$.

The policy $\pi_\theta$ defines the probability of selecting by an agent an action $a$ in state $s$. $\theta$ denotes policy parameters.

\subsection{Pruning}

Pruning is the process of removing unnecessary connections  \cite{lecun1990optimal, hassibi1993surgeon}.
There are many different approaches for finding sparse neural networks and several criterion for classifying algorithms. They could be classified into the following categories: 

\begin{itemize}

\item We fully train a dense model, prune it and finetune \cite{han2015learning}. In this approach, we prune a trained dense network and then finetune remaining weights during additional training steps.
\item Gradually prune dense model during training \cite{zhu_gupta}. Here we start with a dense network and then, according to a specific schedule, which determines the number of weights cut off at each pruning step, we gradually prune the network.
\item Sparse training with a sparse pattern selected \textit{a priori}. In this approach we attempt to prune a dense network at step 0 \cite{lee2018snip, wang2020picking} and keep the topology fixed throughout training. It is worth to note, that training a sparse randomly pruned NN is difficult and leads to much worse results than training a NN with a carefully chosen sparse topology \cite{frankle2018lottery}.
\item Sparse training with a rewiring during training. We start with a sparse NN and maintain sparsity level throughout training, but with the possibility to rewire weights \cite{deep_rewiring, mocanu_set, rigging_lottery}, i.e. to add and to remove connections.

\end{itemize}

On the other side, they could be classified by the pruning criterion. This criterion is used for selecting pruning weights. These criteria are grouped into the Hessian based criteria \cite{lecun1990optimal, hassibi1993surgeon, theis2018faster}, magnitude based \cite{han2015learning} and Bayesian based criteria \cite{molchanov2017variational, dai2018compressing, louizos2017bayesian}. The most widespread in practical applications is the magnitude based approach. In this approach, the smallest by the module weights are pruned.

Also, it is important to distinguish between structured and unstructured pruning. During structural pruning, we remove parameters united in groups (e.g. entire channels, rows, blocks) in order to exploit classical AI hardware efficiently. However, it is important to note that at higher levels of sparsity, structured pruning methods have been observed to lead to a decrease in model accuracy. On the other hand, unstructured pruning does not take any regard to the resulting pattern. This means that parameters are pruned independently, without considering their position or relationship within the model.  Networks pruned with unstructured sparsity usually retain more accuracy compared to structurally pruned counterparts with a similar level of sparsity.

Another important issue is how to distribute pruning weights among layers. There are several approaches:
\begin{itemize}
    \item Global. In this approach, we consider all weights together and select weights for pruning among all weights of the model.
    \item Local uniform. Here we prune in each layer the same fraction of weights.
    \item Local Erdős–Rényi \cite{mocanu_set, rigging_lottery}. Here we make a non-uniform distribution of weights across layers according to the formulae: 

    $s^l = \epsilon * \frac{n^l +n^{l+1}}{n^l * n^{l+1}}$ - for MLP, where $s^l$ is the fraction of the unpruned weights in the layer $l$, $n^l$ is the dimension of the layer $l$, $\epsilon$ is a coefficient for controlling the sparsity level,

    $s^l = \epsilon * \frac{n^l + n^{l+1} + w^l + h^l}{n^l * n^{l+1} * w^l * h^l}$ - for CNNs, where $s^l$ is the fraction of the unpruned weights, $n^l$ is the number of channels in layer $l$, $w^l$ is the convolution kernel width, $h^l$ is the convolution kernel height,  $\epsilon$ is a coefficient for controlling the sparsity level.

    Generally, it is made for reducing the pruning in input and output layers in which usually there is less number of weights due to small input/output dimensions and these layers are more sensible to pruning \cite{han2015learning}.
\end{itemize}


The performance of different pruning techniques for the RL domain was investigated in \cite{rl_sparse}. All training approaches started from sparse NN were usually worse in performance in comparison to the gradual pruning scheme proposed in \cite{zhu_gupta}. 

It was also shown in \cite{rl_sparse} that the performance in almost all MuJoCo environments doesn't degrade even on sparsity levels of 90-95 percents. Another important consequence from \cite{rl_sparse} is that in RL domain sparse NNs could sometimes achieve better performance then their dense counterparts.



\subsection{Quantization}

Generally, quantization is the process of mapping a range of input values to a smaller set of discrete output values. 
Neural network quantization reduces the precision of neural network weights and/or activations. This reduces memory footprint and consequently data transfer 
from memory to processor. Moreover, this enables to use of low-precision/integer arithmetic.
Neural network quantization is a mature field. There are many types of quantization approaches. A comprehensive overview of quantization was presented in \cite{quant_survey}. Here we briefly discuss some important types of quantization. 

Generally, there are two main types of quantization \cite{quant_survey}: 

\begin{itemize}
    \item Quantization aware training (QAT). During training, QAT introduces a non-differentiable quantization operator that quantizes model parameters after each update. However, the weight update and the backward pass are performed in floating point precision. It is crucial to conduct the backward pass using floating point precision as allowing gradient accumulation in quantized precision may lead to zero-gradients or gradients with significant errors, particularly when utilizing low-precision. The reasons for the possibility of using a non-differentiable quantized operator are explained in \cite{ste}. QAT works effectively in practice except for ultra low-precision quantization techniques like binary quantization \cite{quant_survey}.
    \item Post-training quantization (PTQ). An alternative to the QAT is to quantize an already trained model without any fine-tuning. PTQ has a distinct advantage over QAT because it can be used in environments with limited or unlabeled data. Nonetheless, this potentially comes with a cost of decreased accuracy compared to QAT, especially for low-precision quantization techniques.
\end{itemize}

Also, it is necessary to choose a quantization precision. Some methods provide even 1-2 bit precision \cite{binaryconnect, bnns}, however, this usually leads to a strong decrease in accuracy. At the same time, many works show the possibility of using 8-bit precision almost without any decrease in quality.

Moreover, quantization techniques are subdivided by approaches for choosing clipping ranges for weights \cite{quant_survey}:
\begin{itemize}
    \item Quantization could be \textit{symmetric} or \textit{asymmetric}, depending on the symmetry of the clipping interval.
    
    \item \textit{Uniform} and \textit{non-uniform}. In uniform quantization, the input range is divided into equal-sized intervals or steps. In non-uniform the step size is adjusted based on the characteristics of the input signal. Smaller steps are used in regions with more signal activity, and larger steps are used in regions with less activity. Non-uniform quantization may achieve higher accuracy, however it is more complex to implement in hardware.

    \item Quantization granularity. In convolutional layers, different filters could have different ranges of values. This requires to choose the granularity of how the clipping ranges will be calculated. Generally, there are the next approaches: \textit{layerwise (tensorwise)}, \textit{channelwise} and \textit{groupwise}. In layerwise, we calculate one clipping range for all weights in a layer. In channelwise, the quantization is applied independently to each channel within a layer. Channelwise quantization allows for more fine-grained control over the quantization process, considering the characteristics of individual channels. In groupwise quantization, which lay somewhere between the previous two approaches,  channels are grouped together, and quantization is applied to each group.
\end{itemize}

In \cite{quarl} authors analysed both QAT and PTQ 8-bit symmetric quantization for RL tasks. They achieve comparable results with a fully precision training procedure. Moreover, they show that sometimes quantization yields better scores, possibly due to the implicit noise injection during the quantization. 



\section{Methods}

\subsection{RL Algorithms}

For testing optimization algorithm we chose Soft Actor Critic (SAC) \cite{sac} and Deep Q-Network (DQN) \cite{mnih2015human} algorithms due to their popularity and high performance. SAC belongs to the family of actor-critic off-policy algorithms and DQN belongs to the family of value-based off-policy algorithms.

\subsection{RL environments}

We experimented within two RL environments: Mujoco suite  \cite{mujoco_env} and Atari games \cite{atari_env}.

MuJoCo (Multi-Joint dynamics with Contact) environments belong to the class of continuous control environments. Generally, in MuJoCo environments it is necessary to control the behavior (e.g. walking) of biomimetic mechanisms formed within multiple joint rigid bodies. The observations of these environments are vectors of real numbers with dimensions from 8 to 376, that include information about the state of the agent and the world (e.g., positions, velocities, joint angles). The actions (inputs) for these environments are also vectors of real values with dimensions from 1 to 17. They define how the agent can interact with the environment (e.g., applying forces or torques to joints).

Atari environments provide a suite of classic Atari video games as testbeds for reinforcement learning algorithms. Unlike environments that provide low-dimensional state representations, Atari games offer high-dimensional observation spaces directly from the game's pixel output. Actions in Atari games are typically discrete, corresponding to the joystick movements and button presses available on the original Atari 2600 console.

\subsection{Training Procedure}

\begin{figure}
  \includegraphics[width=0.8\linewidth]{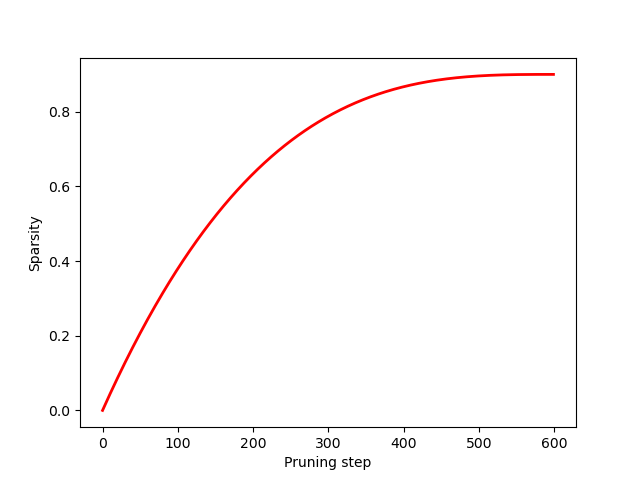}
  \caption{The plot of sparsity function for gradual pruning. The x-axis denotes the pruning step number. The y-axis denotes the neural network sparsity degree.}
  \label{fig:sparsity_plot}
\end{figure}

Since we want to improve the inference, we pruned for SAC only an actor-network. In DQN there is no separation of actor and critic. We start training at environment step $t_0$, then according to \cite{rl_sparse} the pruning commences at environment step $t_s$  and continues until the environment step $t_{f}$. We gradually prune a Neural Network every $\Delta t$ step according to the schedule presented in \cite{zhu_gupta} during $n$ steps.  This pruning scheme involves gradual transformation of a dense network into a sparse one with sparsity $s_{t}$ 
according to formula \ref{sparsity_formula} via weight magnitudes.  When another pruning step is completed, we are leaving the pruned weights equal to zero for the remainder of the training. The training of the pruned NN continues until the environment step $t_p$. The plot of the proposed sparsity schedule is presented in Fig. \ref{fig:sparsity_plot}.


\begin{equation} \label{sparsity_formula}
s_{t} = s_{f} * \left (1 - \left (1 - \frac{t - t_{s}}{n \Delta t} \right )^{3} \right ) for \ t \in \{t_{s}, t_{s} + \Delta t, ... , t_{s} + n \Delta t \}
\end{equation}

For quantizing the pruned NN, after the step $t_p$ we start to apply \textit{symmetric, uniform 8-bit QAT} to the remained weights until the step $t_q$. For fully connected layers we used layerwise quantization. For convolution layers, we used channelwise quantization.




\begin{figure}
  \includegraphics[width=1\linewidth]{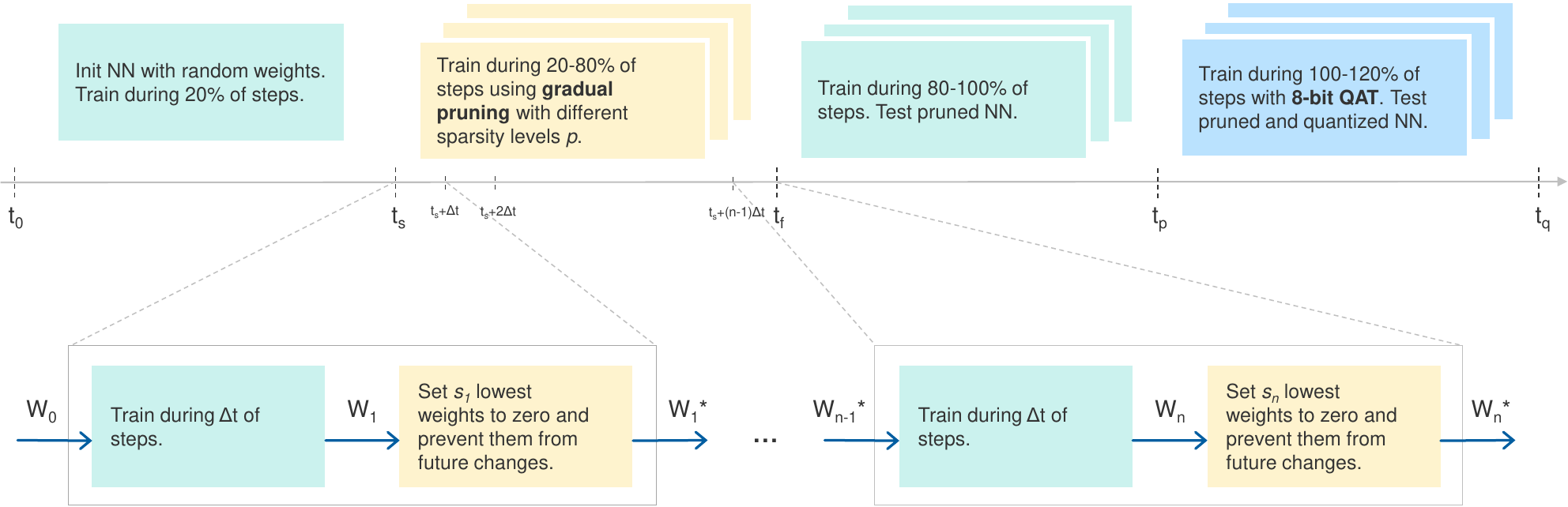}
  \caption{General scheme of training. A randomly initialized neural network is trained for 20\% of the total steps in a classical manner. Further, during the 20-80\% of training, gradual pruning with n steps is applied. Then pruning is turned off and from 80 to 100\% of steps the network is trained again in the classical way. If it is necessary to quantize a NN, additionally 20 \% training steps (step 100-120\%) with 8-bit quantization are performed.}
  \label{fig:training}
\end{figure} 

\section{Experiments}

We experiment within the following RL environments from the MuJoCo suite: HalfCheetah-v4, Hopper-v4, Walker2d-v4, Ant-v4, Humanoid-v4, Swimmer-v4; and Atari games: Pong-v4, Boxing-v4, Tutankham-v4 and CrazyClimber-v4. We repeat each experiment for MuJoCo environments with 10 different seeds. For Atari games, we repeat each experiment with 5 different seeds.

For MuJoCo environments we use the SAC algorithm with a multilayer perceptron (MLP) with two hidden layers with 256 neurons in each of them. The sizes of the input and output layers depend on the environment. All parameters are provided in the Supplementary material.

For Atari environments, we used the DQN algorithm with two different types of neural networks: classical three-layer CNN \cite{mnih2015human} and ResNet \cite{resnet} based networks with three residual blocks \cite{impala}. All parameters are provided in the Supplementary material.

For both algorithms, we used their implementations from the StableBaselines3 \cite{raffin2021stable} library for our experiments.

For each environment, we train sparse policies with different levels of pruning:  50 (x2), 70 (x3.3), 80 (x5), 90 (x10), 95 (x20) and 98 (x50) percent. For MuJoCo environments, we add an additional sparsity level equal to 99 (x100) percent. For each sparsity level we train NN with and without quantization. We start pruning after completing 20 percent ($t_s = 0.2 * total\_steps$)  of steps and finish it after completing 80 percent ($t_f = 0.8 * total\_steps$) of steps. For SAC we use 600 iterations of pruning, and for both DQN we use 300 iterations of pruning. We quantize a neural network after completing the training procedure during an additional 20 percent ($t_q = 1.2 * total\_steps$)  of steps (see Fig. \ref{fig:training}).

In the experimental phase, we employed the Nvidia DGX system. A single experiment, conducted for one environmental setting, required an average of five days of continuous computation for evaluating all possible levels of sparsity, both with and without quantization. In total, the computational duration for all experiments amounted to approximately 40 days.

\section{Results}

\begin{figure}
    \subfigure[Ant-v4]{\includegraphics[width=0.49\textwidth]{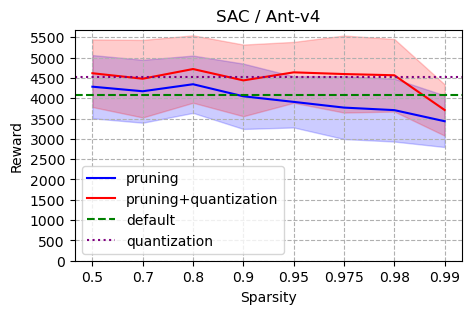}}
    \hfill
    \subfigure[HalfCheetah-v4]{\includegraphics[width=0.49\textwidth]{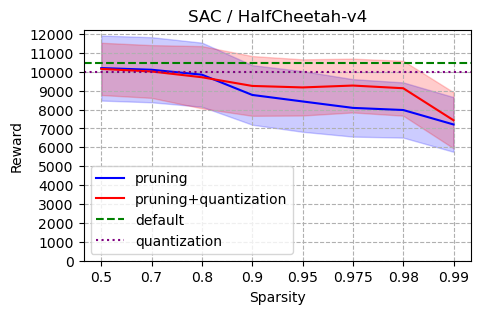}}
    \subfigure[Hopper-v4]{\includegraphics[width=0.49\textwidth]{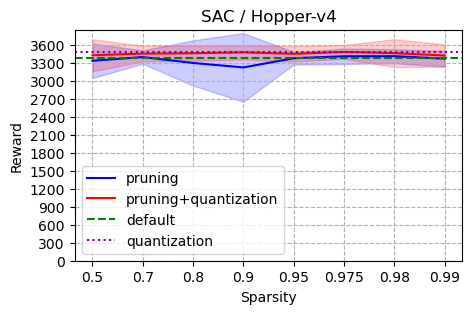}}
    \hfill
    \subfigure[Humanoid-v4]{\includegraphics[width=0.49\textwidth]{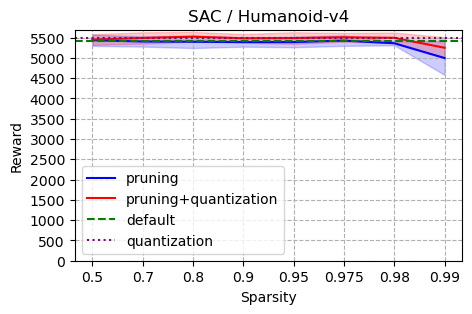}}
    \subfigure[Swimmer-v4]{\includegraphics[width=0.49\textwidth]{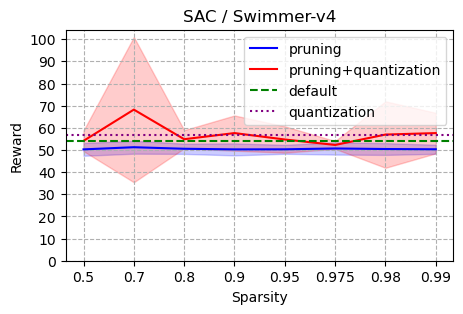}}
    \hfill
    \subfigure[Walker2d-v4]{\includegraphics[width=0.49\textwidth]{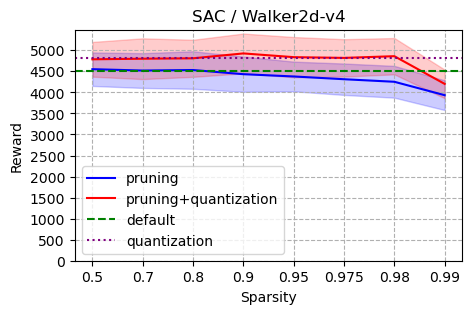}}

\caption{Results for SAC algorithm applied to MuJoCo suite environments. The x-axes of the figures denote the neural network sparsity degree; the y-axes denote the performance -- the reward received by an agent. The blue line shows the performance of the pruned network, and the red line shows the performance of the pruned and quantized network. The dotted purple line shows the performance of the quantized-only network, the green dashed line shows the performance of the default network.
 }
\label{fig:sac_results}
\end{figure}

\begin{figure}
    \subfigure[Pong-v4]{\includegraphics[width=0.49\textwidth]{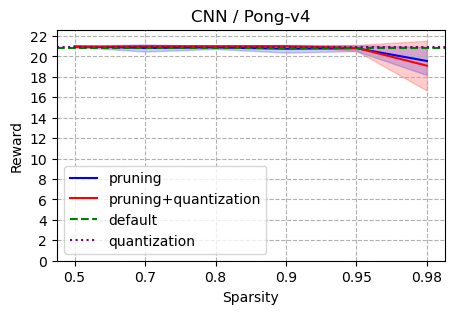}}
    \hfill
    \subfigure[Enduro-v4]{\includegraphics[width=0.49\textwidth]{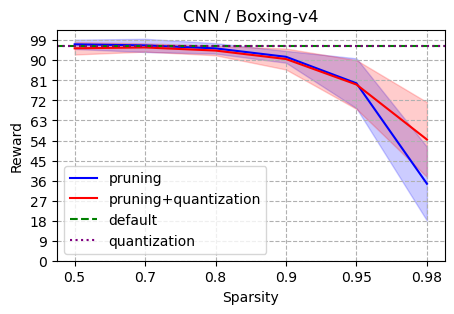}}
    \subfigure[Tutankham-v4]{\includegraphics[width=0.49\textwidth]{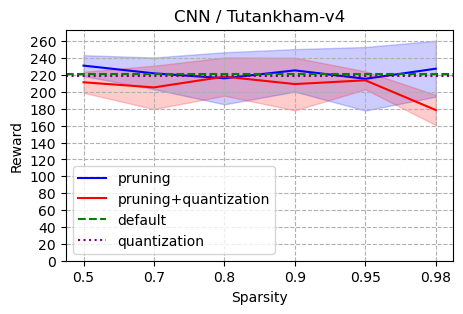}}
    \hfill
    \subfigure[CrazyClimber-v4]{\includegraphics[width=0.49\textwidth] {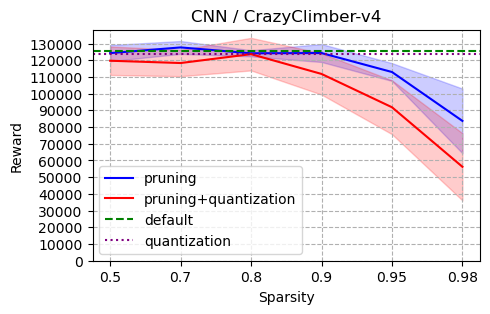}}

\caption{Results for DQN algorithm based on the CNN applied to Atari environments. The x-axes of the figures denote the neural network sparsity degree; the y-axes denote the performance -- the reward received by an agent. The blue line shows the performance of the pruned network and the red line shows the performance of the pruned and quantized network. The dotted purple line shows the performance of the quantized-only network, green dashed line shows the performance of the default dense and fully precision network.
 }
\label{fig:dqn_results}
\end{figure}

\begin{figure}
    \subfigure[Pong-v4]{\includegraphics[width=0.49\textwidth]{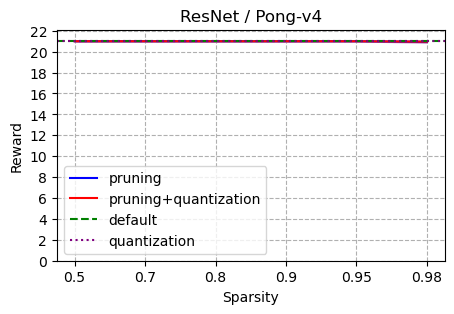}}
    \hfill
    \subfigure[CrazyClimber-v4]{\includegraphics[width=0.49\textwidth]{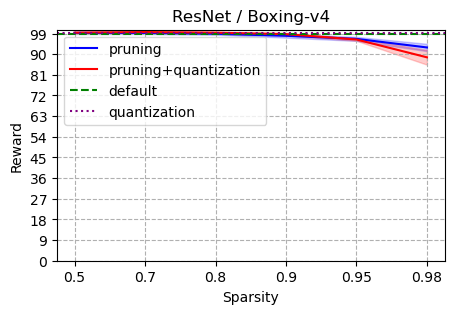}}
    \subfigure[Pong-v4]{\includegraphics[width=0.49\textwidth]{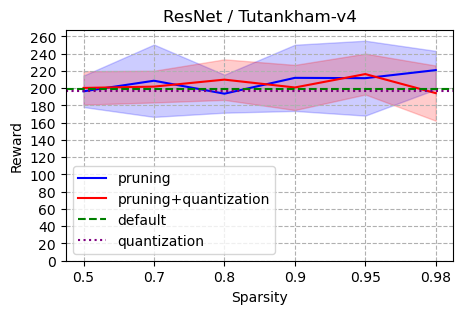}}
    \hfill
    \subfigure[Pong-v4]{\includegraphics[width=0.49\textwidth]{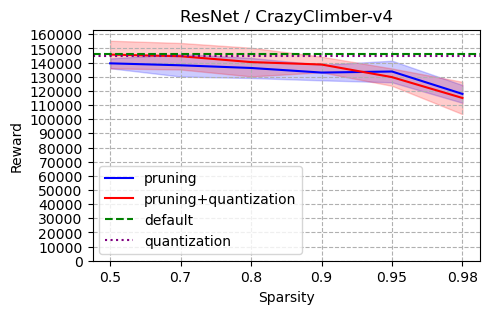}}

\caption{Results for DQN algorithm based on the ResNet applied to Atari environments. The x-axes of the figures denote the neural network sparsity degree; the y-axes denote the performance -- the reward received by an agent. The blue line shows the performance of the pruned network, and the red line shows the performance of the pruned and quantized network. The dotted purple line shows the performance of the quantized only network, green dashed line shows the performance of the default dense and fully precision network.
 }
\label{fig:dqn_resnet_results}
\end{figure}

Figures \ref{fig:sac_results}, \ref{fig:dqn_results}, \ref{fig:dqn_resnet_results} present the performance of pruned and/or quantized neural networks in various environments.

We see in Fig. \ref{fig:sac_results} that for the most number of MuJoCo environments (except HalfCheetah) we could prune and quantize up to 98 percent without the loss in quality, which leads to a 200x decrease in the size of neural networks: 4x by quantization, 50x by pruning. Even for HalfCheetah we could prune 80 percent of weights and quantize them, which leads to a 20x decrease in the size of the neural network. For some environments e.g. Hopper and Swimmer we could prune 99 percent of weights and quantize them without the loss in quality which leads to a 400x decrease in the size of the neural network. Furthermore, quantization+pruning usually slightly outperforms pruning, which leads to better results even in comparison to the dense model.

For classical CNN-based DQN for Atari environments, we see in Fig. \ref{fig:dqn_results} that for all environments we could prune and quantize up to 80 percent without the loss in quality, which leads to a 20x decrease in the of optimized neural networks. For Pong and Tutankham we could prune and quantize up to 95 percent of sparsity which leads to a total 100x decrease in the size of neural networks.

For ResNet-based DQN for Atari environments, we see in Fig. \ref{fig:dqn_resnet_results} the possibility to prune and quantize up to 95 percent, without the significant loss in quality, that leads to a 80x decrease in the size of neural networks. For Pong and Tutankham we could prune and quantize up to 98 percent of sparsity which leads to a 200x decrease in the size of neural networks. It is worth noting that ResNet-based networks are much more suitable for pruning and quantizing which coincide with the findings in \cite{rl_sparse}.





\section{Discussion}

In this study, we demonstrated the large redundancy (up to 400x) in the neural network size used for popular RL tasks. In some sense, quantization and pruning could be considered as neuromorphic approaches. In the brain, there are no fully connected layers \cite{principles_of_ns} and strong regular structure in comparison to modern NNs. Also, it seems impossible to store values with the precision provided by the 32-bit floating points in highly noisy cell environment \cite{brain_discrete, perception_discrete, quant_survey}.

Minimizing the size of NNs mitigates the von Neumann problem of modern hardware, by reducing the exchange between memory and processor. Moreover, often it is possible to locate obtained smallified NNs in on-chip memory. That could lead to very high inference speeds, low energy consumption and low latencies. It was shown that this desire could be achieved even on classical CPUs by the Neural Magic company for classical DL domains. Moreover, the recent IBM chip NorthPole \cite{modha2023ibm} based totally on near-memory computing and storing weights and activations in the on-chip memory, could be enhanced by optimization algorithms proposed here.

By providing the possibility of significantly reducing neural networks trained by RL algorithms, we provide more possibilities for their use in practical areas such as Edge AI, real-time control, robotics, and many others. But it is worth to note, that the maximum profit could be achieved in a smart co-design of algorithms and hardware.

\section{Author contributions}
DI, DL, OM, and VV contributed to the conception and design of the study. DI and DL were contributed equally. OM, and VV were co-senior authors. All authors contributed to manuscript revision, read, and approved the submitted version.

\section{Acknowledgments}
The research is carried out using the equipment of the shared research facilities of HPC computing resources at Lomonosov Moscow State University \cite{voevodin2019supercomputer} and Cifrum IT infrastructure. The work was supported by the Russian Science Foundation, project 23-72-10088.




\bibliographystyle{unsrtnat}
\bibliography{references}  

\begin{thebibliography}{48}
\providecommand{\natexlab}[1]{#1}
\providecommand{\url}[1]{\texttt{#1}}
\expandafter\ifx\csname urlstyle\endcsname\relax
  \providecommand{\doi}[1]{doi: #1}\else
  \providecommand{\doi}{doi: \begingroup \urlstyle{rm}\Url}\fi

\bibitem[Mnih et~al.(2015)Mnih, Kavukcuoglu, Silver, Rusu, Veness, Bellemare,
  Graves, Riedmiller, Fidjeland, Ostrovski, et~al.]{mnih2015human}
Volodymyr Mnih, Koray Kavukcuoglu, David Silver, Andrei~A Rusu, Joel Veness,
  Marc~G Bellemare, Alex Graves, Martin Riedmiller, Andreas~K Fidjeland, Georg
  Ostrovski, et~al.
\newblock Human-level control through deep reinforcement learning.
\newblock \emph{nature}, 518\penalty0 (7540):\penalty0 529--533, 2015.

\bibitem[Degrave et~al.(2022)Degrave, Felici, Buchli, Neunert, Tracey,
  Carpanese, Ewalds, Hafner, Abdolmaleki, de~Las~Casas, et~al.]{rl_tokamak}
Jonas Degrave, Federico Felici, Jonas Buchli, Michael Neunert, Brendan Tracey,
  Francesco Carpanese, Timo Ewalds, Roland Hafner, Abbas Abdolmaleki, Diego
  de~Las~Casas, et~al.
\newblock Magnetic control of tokamak plasmas through deep reinforcement
  learning.
\newblock \emph{Nature}, 602\penalty0 (7897):\penalty0 414--419, 2022.

\bibitem[Kaufmann et~al.(2023)Kaufmann, Bauersfeld, Loquercio, M{\"u}ller,
  Koltun, and Scaramuzza]{rl_drone}
Elia Kaufmann, Leonard Bauersfeld, Antonio Loquercio, Matthias M{\"u}ller,
  Vladlen Koltun, and Davide Scaramuzza.
\newblock Champion-level drone racing using deep reinforcement learning.
\newblock \emph{Nature}, 620\penalty0 (7976):\penalty0 982--987, 2023.

\bibitem[Gholami et~al.(2021)Gholami, Kim, Dong, Yao, Mahoney, and
  Keutzer]{quant_survey}
Amir Gholami, Sehoon Kim, Zhen Dong, Zhewei Yao, Michael~W Mahoney, and Kurt
  Keutzer.
\newblock A survey of quantization methods for efficient neural network
  inference.
\newblock \emph{arXiv preprint arXiv:2103.13630}, 2021.

\bibitem[Liang et~al.(2021)Liang, Glossner, Wang, Shi, and
  Zhang]{liang2021pruning}
Tailin Liang, John Glossner, Lei Wang, Shaobo Shi, and Xiaotong Zhang.
\newblock Pruning and quantization for deep neural network acceleration: A
  survey.
\newblock \emph{Neurocomputing}, 461:\penalty0 370--403, 2021.

\bibitem[Yousefzadeh et~al.(2019)Yousefzadeh, Khoei, Hosseini, Holanda, Leroux,
  Moreira, Tapson, Dhoedt, Simoens, Serrano-Gotarredona,
  et~al.]{yousefzadeh2019asynchronous}
Amirreza Yousefzadeh, Mina~A Khoei, Sahar Hosseini, Priscila Holanda, Sam
  Leroux, Orlando Moreira, Jonathan Tapson, Bart Dhoedt, Pieter Simoens, Teresa
  Serrano-Gotarredona, et~al.
\newblock Asynchronous spiking neurons, the natural key to exploit temporal
  sparsity.
\newblock \emph{IEEE Journal on Emerging and Selected Topics in Circuits and
  Systems}, 9\penalty0 (4):\penalty0 668--678, 2019.

\bibitem[Ivanov et~al.(2023)Ivanov, Larionov, Kiselev, and
  Dylov]{significant_multiplications}
Dmitry~A Ivanov, Denis~A Larionov, Mikhail~V Kiselev, and Dmitry~V Dylov.
\newblock Deep reinforcement learning with significant multiplications
  inference.
\newblock \emph{Scientific Reports}, 13\penalty0 (1):\penalty0 20865, 2023.

\bibitem[Hinton et~al.(2015)Hinton, Vinyals, and Dean]{hinton2015distilling}
Geoffrey Hinton, Oriol Vinyals, and Jeff Dean.
\newblock Distilling the knowledge in a neural network.
\newblock \emph{arXiv preprint arXiv:1503.02531}, 2015.

\bibitem[Ren et~al.(2021)Ren, Xiao, Chang, Huang, Li, Chen, and
  Wang]{ren2021comprehensive}
Pengzhen Ren, Yun Xiao, Xiaojun Chang, Po-Yao Huang, Zhihui Li, Xiaojiang Chen,
  and Xin Wang.
\newblock A comprehensive survey of neural architecture search: Challenges and
  solutions.
\newblock \emph{ACM Computing Surveys (CSUR)}, 54\penalty0 (4):\penalty0 1--34,
  2021.

\bibitem[Han et~al.(2016)Han, Liu, Mao, Pu, Pedram, Horowitz, and Dally]{eie}
Song Han, Xingyu Liu, Huizi Mao, Jing Pu, Ardavan Pedram, Mark~A Horowitz, and
  William~J Dally.
\newblock Eie: Efficient inference engine on compressed deep neural network.
\newblock \emph{ACM SIGARCH Computer Architecture News}, 44\penalty0
  (3):\penalty0 243--254, 2016.

\bibitem[Han et~al.(2015{\natexlab{a}})Han, Mao, and Dally]{deep_compression}
Song Han, Huizi Mao, and William~J Dally.
\newblock Deep compression: Compressing deep neural networks with pruning,
  trained quantization and huffman coding.
\newblock \emph{arXiv preprint arXiv:1510.00149}, 2015{\natexlab{a}}.

\bibitem[Blalock et~al.(2020)Blalock, Ortiz, Frankle, and
  Guttag]{blalock2020state}
Davis Blalock, Jose Javier~Gonzalez Ortiz, Jonathan Frankle, and John Guttag.
\newblock What is the state of neural network pruning?
\newblock \emph{arXiv preprint arXiv:2003.03033}, 2020.

\bibitem[Graesser et~al.(2022)Graesser, Evci, Elsen, and Castro]{rl_sparse}
Laura Graesser, Utku Evci, Erich Elsen, and Pablo~Samuel Castro.
\newblock The state of sparse training in deep reinforcement learning.
\newblock In \emph{International Conference on Machine Learning}, pages
  7766--7792. PMLR, 2022.

\bibitem[Frankle and Carbin(2018)]{frankle2018lottery}
Jonathan Frankle and Michael Carbin.
\newblock The lottery ticket hypothesis: Finding sparse, trainable neural
  networks.
\newblock \emph{arXiv preprint arXiv:1803.03635}, 2018.

\bibitem[Hudspeth et~al.(2013)Hudspeth, Jessell, Kandel, Schwartz, and
  Siegelbaum]{principles_of_ns}
A~James Hudspeth, Thomas~M Jessell, Eric~R Kandel, James~Harris Schwartz, and
  Steven~A Siegelbaum.
\newblock \emph{Principles of neural science}.
\newblock McGraw-Hill, Health Professions Division, 2013.

\bibitem[Tee and Taylor(2020)]{brain_discrete}
James Tee and Desmond~P Taylor.
\newblock Is information in the brain represented in continuous or discrete
  form?
\newblock \emph{IEEE Transactions on Molecular, Biological and Multi-Scale
  Communications}, 6\penalty0 (3):\penalty0 199--209, 2020.

\bibitem[VanRullen and Koch(2003)]{perception_discrete}
Rufin VanRullen and Christof Koch.
\newblock Is perception discrete or continuous?
\newblock \emph{Trends in cognitive sciences}, 7\penalty0 (5):\penalty0
  207--213, 2003.

\bibitem[Faisal et~al.(2008)Faisal, Selen, and Wolpert]{noise_nervous_system}
A~Aldo Faisal, Luc~PJ Selen, and Daniel~M Wolpert.
\newblock Noise in the nervous system.
\newblock \emph{Nature reviews neuroscience}, 9\penalty0 (4):\penalty0
  292--303, 2008.

\bibitem[Sun et~al.(2012)Sun, Wang, Goyal, and Varshney]{bayesian_framework}
John~Z Sun, Grace~I Wang, Vivek~K Goyal, and Lav~R Varshney.
\newblock A framework for bayesian optimality of psychophysical laws.
\newblock \emph{Journal of Mathematical Psychology}, 56\penalty0 (6):\penalty0
  495--501, 2012.

\bibitem[Horowitz(2014)]{horowitz2014}
Mark Horowitz.
\newblock 1.1 computing's energy problem (and what we can do about it).
\newblock In \emph{2014 IEEE International Solid-State Circuits Conference
  Digest of Technical Papers (ISSCC)}, pages 10--14. IEEE, 2014.

\bibitem[Backus(1978)]{backus1978can}
John Backus.
\newblock Can programming be liberated from the von neumann style? a functional
  style and its algebra of programs.
\newblock \emph{Communications of the ACM}, 21\penalty0 (8):\penalty0 613--641,
  1978.

\bibitem[Ivanov et~al.(2022)Ivanov, Chezhegov, Kiselev, Grunin, and
  Larionov]{neuromorphic_systems}
Dmitry Ivanov, Aleksandr Chezhegov, Mikhail Kiselev, Andrey Grunin, and Denis
  Larionov.
\newblock Neuromorphic artificial intelligence systems.
\newblock \emph{Frontiers in Neuroscience}, 16, 2022.

\bibitem[Krishnan et~al.(2019)Krishnan, Lam, Chitlangia, Wan, Barth-Maron,
  Faust, and Reddi]{quarl}
Srivatsan Krishnan, Maximilian Lam, Sharad Chitlangia, Zishen Wan, Gabriel
  Barth-Maron, Aleksandra Faust, and Vijay~Janapa Reddi.
\newblock Quarl: Quantization for fast and environmentally sustainable
  reinforcement learning.
\newblock \emph{arXiv preprint arXiv:1910.01055}, 2019.

\bibitem[Sutton and Barto(2018)]{sutton2018reinforcement}
Richard~S Sutton and Andrew~G Barto.
\newblock \emph{Reinforcement learning: An introduction}.
\newblock MIT press, 2018.

\bibitem[LeCun et~al.(1990)LeCun, Denker, and Solla]{lecun1990optimal}
Yann LeCun, John~S Denker, and Sara~A Solla.
\newblock Optimal brain damage.
\newblock In \emph{Advances in neural information processing systems}, pages
  598--605, 1990.

\bibitem[Hassibi and Stork(1993)]{hassibi1993surgeon}
Babak Hassibi and David~G Stork.
\newblock \emph{Second order derivatives for network pruning: Optimal brain
  surgeon}.
\newblock Morgan Kaufmann, 1993.

\bibitem[Han et~al.(2015{\natexlab{b}})Han, Pool, Tran, and
  Dally]{han2015learning}
Song Han, Jeff Pool, John Tran, and William Dally.
\newblock Learning both weights and connections for efficient neural network.
\newblock \emph{Advances in neural information processing systems}, 28,
  2015{\natexlab{b}}.

\bibitem[Zhu and Gupta(2017)]{zhu_gupta}
Michael Zhu and Suyog Gupta.
\newblock To prune, or not to prune: exploring the efficacy of pruning for
  model compression.
\newblock \emph{arXiv preprint arXiv:1710.01878}, 2017.

\bibitem[Lee et~al.(2018)Lee, Ajanthan, and Torr]{lee2018snip}
Namhoon Lee, Thalaiyasingam Ajanthan, and Philip~HS Torr.
\newblock Snip: Single-shot network pruning based on connection sensitivity.
\newblock \emph{arXiv preprint arXiv:1810.02340}, 2018.

\bibitem[Wang et~al.(2020)Wang, Zhang, and Grosse]{wang2020picking}
Chaoqi Wang, Guodong Zhang, and Roger Grosse.
\newblock Picking winning tickets before training by preserving gradient flow.
\newblock \emph{arXiv preprint arXiv:2002.07376}, 2020.

\bibitem[Bellec et~al.(2017)Bellec, Kappel, Maass, and
  Legenstein]{deep_rewiring}
Guillaume Bellec, David Kappel, Wolfgang Maass, and Robert Legenstein.
\newblock Deep rewiring: Training very sparse deep networks.
\newblock \emph{arXiv preprint arXiv:1711.05136}, 2017.

\bibitem[Mocanu et~al.(2018)Mocanu, Mocanu, Stone, Nguyen, Gibescu, and
  Liotta]{mocanu_set}
Decebal~Constantin Mocanu, Elena Mocanu, Peter Stone, Phuong~H Nguyen,
  Madeleine Gibescu, and Antonio Liotta.
\newblock Scalable training of artificial neural networks with adaptive sparse
  connectivity inspired by network science.
\newblock \emph{Nature communications}, 9\penalty0 (1):\penalty0 2383, 2018.

\bibitem[Evci et~al.(2020)Evci, Gale, Menick, Castro, and
  Elsen]{rigging_lottery}
Utku Evci, Trevor Gale, Jacob Menick, Pablo~Samuel Castro, and Erich Elsen.
\newblock Rigging the lottery: Making all tickets winners.
\newblock In \emph{International Conference on Machine Learning}, pages
  2943--2952. PMLR, 2020.

\bibitem[Theis et~al.(2018)Theis, Korshunova, Tejani, and
  Husz{\'a}r]{theis2018faster}
Lucas Theis, Iryna Korshunova, Alykhan Tejani, and Ferenc Husz{\'a}r.
\newblock Faster gaze prediction with dense networks and fisher pruning.
\newblock \emph{arXiv preprint arXiv:1801.05787}, 2018.

\bibitem[Molchanov et~al.(2017)Molchanov, Ashukha, and
  Vetrov]{molchanov2017variational}
Dmitry Molchanov, Arsenii Ashukha, and Dmitry Vetrov.
\newblock Variational dropout sparsifies deep neural networks.
\newblock In \emph{International Conference on Machine Learning}, pages
  2498--2507. PMLR, 2017.

\bibitem[Dai et~al.(2018)Dai, Zhu, Guo, and Wipf]{dai2018compressing}
Bin Dai, Chen Zhu, Baining Guo, and David Wipf.
\newblock Compressing neural networks using the variational information
  bottleneck.
\newblock In \emph{International Conference on Machine Learning}, pages
  1135--1144. PMLR, 2018.

\bibitem[Louizos et~al.(2017)Louizos, Ullrich, and
  Welling]{louizos2017bayesian}
Christos Louizos, Karen Ullrich, and Max Welling.
\newblock Bayesian compression for deep learning.
\newblock \emph{Advances in neural information processing systems}, 30, 2017.

\bibitem[Yin et~al.(2019)Yin, Lyu, Zhang, Osher, Qi, and Xin]{ste}
Penghang Yin, Jiancheng Lyu, Shuai Zhang, Stanley Osher, Yingyong Qi, and Jack
  Xin.
\newblock Understanding straight-through estimator in training activation
  quantized neural nets.
\newblock \emph{arXiv preprint arXiv:1903.05662}, 2019.

\bibitem[Courbariaux et~al.(2015)Courbariaux, Bengio, and David]{binaryconnect}
Matthieu Courbariaux, Yoshua Bengio, and Jean-Pierre David.
\newblock Binaryconnect: Training deep neural networks with binary weights
  during propagations.
\newblock \emph{Advances in neural information processing systems}, 28, 2015.

\bibitem[Hubara et~al.(2016)Hubara, Courbariaux, Soudry, El-Yaniv, and
  Bengio]{bnns}
Itay Hubara, Matthieu Courbariaux, Daniel Soudry, Ran El-Yaniv, and Yoshua
  Bengio.
\newblock Binarized neural networks.
\newblock \emph{Advances in neural information processing systems}, 29, 2016.

\bibitem[Haarnoja et~al.(2018)Haarnoja, Zhou, Abbeel, and Levine]{sac}
Tuomas Haarnoja, Aurick Zhou, Pieter Abbeel, and Sergey Levine.
\newblock Soft actor-critic: Off-policy maximum entropy deep reinforcement
  learning with a stochastic actor.
\newblock In \emph{International conference on machine learning}, pages
  1861--1870. PMLR, 2018.

\bibitem[Todorov et~al.(2012)Todorov, Erez, and Tassa]{mujoco_env}
Emanuel Todorov, Tom Erez, and Yuval Tassa.
\newblock Mujoco: A physics engine for model-based control.
\newblock In \emph{2012 IEEE/RSJ international conference on intelligent robots
  and systems}, pages 5026--5033. IEEE, 2012.

\bibitem[Bellemare et~al.(2012)Bellemare, Veness, and Bowling]{atari_env}
Marc Bellemare, Joel Veness, and Michael Bowling.
\newblock Investigating contingency awareness using atari 2600 games.
\newblock In \emph{Proceedings of the AAAI Conference on Artificial
  Intelligence}, volume~26, pages 864--871, 2012.

\bibitem[He et~al.(2016)He, Zhang, Ren, and Sun]{resnet}
Kaiming He, Xiangyu Zhang, Shaoqing Ren, and Jian Sun.
\newblock Deep residual learning for image recognition.
\newblock In \emph{Proceedings of the IEEE conference on computer vision and
  pattern recognition}, pages 770--778, 2016.

\bibitem[Espeholt et~al.(2018)Espeholt, Soyer, Munos, Simonyan, Mnih, Ward,
  Doron, Firoiu, Harley, Dunning, et~al.]{impala}
Lasse Espeholt, Hubert Soyer, Remi Munos, Karen Simonyan, Vlad Mnih, Tom Ward,
  Yotam Doron, Vlad Firoiu, Tim Harley, Iain Dunning, et~al.
\newblock Impala: Scalable distributed deep-rl with importance weighted
  actor-learner architectures.
\newblock In \emph{International conference on machine learning}, pages
  1407--1416. PMLR, 2018.

\bibitem[Raffin et~al.(2021)Raffin, Hill, Gleave, Kanervisto, Ernestus, and
  Dormann]{raffin2021stable}
Antonin Raffin, Ashley Hill, Adam Gleave, Anssi Kanervisto, Maximilian
  Ernestus, and Noah Dormann.
\newblock Stable-baselines3: Reliable reinforcement learning implementations.
\newblock \emph{Journal of Machine Learning Research}, 22\penalty0
  (268):\penalty0 1--8, 2021.

\bibitem[Modha et~al.(2023)Modha, Akopyan, Andreopoulos, Appuswamy, Arthur,
  Cassidy, Datta, DeBole, Esser, Otero, et~al.]{modha2023ibm}
Dharmendra~S Modha, Filipp Akopyan, Alexander Andreopoulos, Rathinakumar
  Appuswamy, John~V Arthur, Andrew~S Cassidy, Pallab Datta, Michael~V DeBole,
  Steven~K Esser, Carlos~Ortega Otero, et~al.
\newblock Ibm northpole neural inference machine.
\newblock In \emph{2023 IEEE Hot Chips 35 Symposium (HCS)}, pages 1--58. IEEE
  Computer Society, 2023.

\bibitem[Voevodin et~al.(2019)Voevodin, Antonov, Nikitenko, Shvets, Sobolev,
  Sidorov, Stefanov, Voevodin, and Zhumatiy]{voevodin2019supercomputer}
Vladimir~V Voevodin, Alexander~S Antonov, Dmitry~A Nikitenko, Pavel~A Shvets,
  Sergey~I Sobolev, Igor~Yu Sidorov, Konstantin~S Stefanov, Vadim~V Voevodin,
  and Sergey~A Zhumatiy.
\newblock Supercomputer lomonosov-2: Large scale, deep monitoring and fine
  analytics for the user community.
\newblock \emph{Supercomputing Frontiers and Innovations}, 6\penalty0
  (2):\penalty0 4--11, 2019.

\end{thebibliography}

\section{Supplementary}

\subsection{Experimental details}

\begin{center}
\begin{table}[h!]
\centering
\begin{tabular}{ c|c } 
 \hline
 Parameter & Value \\
 \hline
 optimizer & Adam \\
 learning rate & $1 * 10^{-4}$ \\
 Adam epsilon & $1 * 10^{-8}$ \\
 weight decay & $0$ \\
 discount $\gamma$ & $0.99$ \\
 replay buffer size & $10^6$ \\
 target update interval & $8,000$  \\
 target smoothing coefficient ($\tau$) & 1.0 \\
 train frequency & $4$ \\
 gradient steps & $1$ \\
 batch size & $32$ \\
 learning starts (initial collect steps) & $20,000$  \\
 \hline
 exploration fraction & $0.01$  \\
 exploration initial epsilon & $1.0$ \\
 exploration final epsilon & $0.01$ \\

 \hline
 hidden CNN layers & $3$ \\
 layer 1 (filters, kernel, stride) & 32, 8, 4 \\
 layer 2 (filters, kernel, stride) & 64, 4, 2 \\
 layer 3 (filters, kernel, stride) & 64, 3, 1 \\
 hidden dense layers & $1$ \\
 neurons per hidden layer & $512$ \\
 nonlinearity & ReLU \\
 \hline
 training episodes & $10,000,000$ \\
 pruning interval & $20,000$ \\
 \hline
 evaluation frequency & $100,000$ \\
 evaluation episodes & $10$ \\
 \hline
 
\end{tabular}
\caption{DQN CNN hyperparameters}
\label{table:1}
\end{table}
\end{center}

\begin{center}
\begin{table}[h!]
\centering
\begin{tabular}{ c|c } 
 \hline
 Parameter & Value \\
 \hline
 optimizer & Adam \\
 learning rate & $1 * 10^{-4}$ \\
 Adam epsilon & $3.125 * 10^{-4}$ \\
 weight decay & $1 * 10^{-5}$ \\
 discount $\gamma$ & $0.99$ \\
 replay buffer size & $10^6$ \\
 target update interval & $8,000$ \\
 target smoothing coefficient ($\tau$) & 1.0 \\
 train frequency & $4$ \\
 gradient steps & $1$ \\
 batch size & $32$ \\
 learning starts (initial collect steps) & $20,000$ \\
 \hline
 exploration fraction & $0.01$ \\
 exploration initial epsilon & $1.0$ \\
 exploration final epsilon & $0.01$ \\

 \hline
 \textit{ResNet Architecture:} & \\
 number of stacks & $3$ \\
 hidden dense layers & $1$ \\
 neurons per hidden layer & $512$ \\
 nonlinearity & ReLU \\
 \hdashline
 \textit{ResNet stack block:} & \\
 CNN layers & 1 \\
 max pool layers & 1 \\
 residual-CNN layers & 2 \\
 \hdashline
 \textit{ResNet stack blocks params:} & \\
 stack 1 (filters, kernel, stride) & 32, 8, 4 \\
 stack 2 (filters, kernel, stride) & 64, 4, 2 \\
 stack 3 (filters, kernel, stride) & 64, 3, 1 \\
 \hline
 training episodes & $10,000,000$ \\
 pruning interval & $20,000$ \\
 \hline
 evaluation frequency & $100,000$ \\
 evaluation episodes & $10$ \\
 \hline
\end{tabular}
\caption{DQN ResNet hyperparameters}
\label{table:2}
\end{table}
\end{center}

\begin{center}
\begin{table}[h!]
\centering
\begin{tabular}{ c|c } 
 \hline
 Parameter & Value \\
 \hline
 optimizer & Adam \\
 learning rate & $3 * 10^{-4}$ \\
 weight decay & $1 * 10^{-4}$ \\
 discount & $0.99$ \\
 replay buffer size & $10^6$ \\
 target update interval & $1$ \\
 target smoothing coefficient ($\tau$) & 0.005 \\
 train frequency & $1$ \\
 gradient steps & $1$ \\
 batch size & $256$ \\
 learning starts (initial collect steps) & - \\
 
 \hline
 hidden layers & $2$ \\
 neurons per hidden layer & $256$ \\
 nonlinearity & ReLU \\
 \hline
 training episodes & $1,000,000$ \\
 pruning interval & $1,000$ \\
 \hline
 evaluation frequency & $10,000$ \\
 evaluation episodes & $20$ \\
 \hline
\end{tabular}
\caption{SAC hyperparameters}
\label{table:3}
\end{table}
\end{center}

\end{document}